# PetroBench: A Benchmark for Large Language Models in Petroleum Engineering

Xiang Wang [a, *], Tingting Zhang [a], Sen Wang [b], Ying Wu [a], Heng Meng [a], Peng Zhou [a], Peng Li [a]

[a] School of Petroleum and Natural Gas Engineering, Changzhou University, Changzhou 213164, China
[b] China University of Petroleum (East China), Qingdao, Shandong 266580, China

**Abstract:** Large Language Models (LLMs) are experiencing rapid growth, accompanied by increasing applications in the petroleum industry. Given the diversity of available models and the complexity of engineering scenarios, the industry urgently requires a scientific evaluation framework tailored to petroleum engineering to support model selection and optimization. To address this need, a domain-specific evaluation benchmark system for LLMs in petroleum engineering was developed, providing a unified and quantifiable basis for model selection, performance comparison, and continuous improvement. A three-stage evaluation and screening framework was designed, comprising data preprocessing, quality assessment and filtering, and multi-model response validation. A question quality analysis matrix and a multidimensional weighted scoring mechanism were established, complemented by expert review, to construct a standardized question bank with high domain relevance and discriminative capability. The benchmark covers three core disciplines—production engineering, reservoir engineering, and drilling engineering—and includes four question types: multiple-choice questions, true/false questions, term definition, and short-answer questions, totaling 1,200 items. Eight mainstream models (Qwen3-Max, Gemini-3-Pro, Grok-4.2, Claude-Opus-4.6-Thinking, DeepSeek-V3.2-Thinking, Doubao-Seed-1.8-Thinking, Kimi-K2.5, and GPT-5.4) were evaluated under a unified application programming interface (API) environment. The results indicate that model performance on subjective questions exceeded that on objective questions, revealing limitations in precise discrimination of factual knowledge in petroleum engineering. The highest accuracy rates for multiple-choice and true/false questions were 65.3% (Kimi-K2.5) and 74.3% (Gemini-3-Pro), respectively. Overall scores exhibited a three-tier distribution, with Gemini-3-Pro, Kimi-K2.5, and Claude-Opus-4.6-Thinking leading at 72%–74%. Discipline-specific analysis showed the best performance in production engineering, comparatively lower scores in reservoir engineering, and the most pronounced inter-type performance variation in drilling engineering. Comparative analysis revealed that international models slightly outperformed Chinese models in short-answer questions (+0.55 points), whereas Chinese models demonstrated higher accuracy in multiple-choice questions (+3.6 percentage points), with comparable performance in true/false questions. In addition, a trade-off between accuracy and response latency was observed, as higher-accuracy models tended to exhibit longer response times. Overall, the proposed benchmark demonstrates strong domain specificity, discriminative capability, and reproducibility. It provides a standardized reference for the scientific evaluation and engineering deployment of LLMs in the petroleum industry, with substantial potential for broader application.

# 1 Introduction

## 1.1 Research Background

In recent years, Large Language Models (LLMs) have evolved rapidly. From the GPT series to Claude and Gemini, continuous growth in parameter scale has substantially enhanced their capabilities in natural language understanding, logical reasoning, and knowledge generation[1,2], leading to strong overall performance in open-domain tasks. At present, two primary pathways are commonly adopted for industrial deployment of LLMs. The first involves direct utilization of general-purpose LLMs to solve domain-specific tasks through prompting and in-context learning [1,3]. The second involves construction of domain-specific models based on general foundation models through transfer learning, instruction tuning, fine-tuning, and parameter-efficient adaptation techniques [4–7]. These approaches have significantly reduced the cost of industrial AI deployment while improving adaptability to specialized knowledge and engineering scenarios.

Petroleum engineering is characterized by significant technical barriers and high engineering complexity. Its business scope spans exploration, drilling, production, and reservoir management, involving interdisciplinary integration across geology, fluid mechanics, and materials science, while operating under stringent safety and environmental regulations. Problems in professional scenarios typically exhibit strong engineering logic, high risk sensitivity, and explicit task orientation.

In response to these characteristics, the petroleum industry has initiated vertical exploration of LLM applications. In China, Sinopec Shengli Oilfield has introduced the "Sheng Xiaoli" model, while the PetroChina Research Institute of Petroleum Exploration and Development has developed PetroAI and collaborated with technology enterprises to construct industry-oriented models for pipeline design and full-process operations management. Internationally, major companies such as ExxonMobil and Shell have begun integrating generative artificial intelligence (AI) into geological modeling and asset management workflows. These developments indicate a growing consensus that LLMs can enhance operational efficiency in the petroleum sector.

Nevertheless, regardless of the deployment pathway, a critical challenge persists: the absence of a unified and standardized evaluation framework tailored to petroleum engineering tasks. On the one hand, users lack scientific criteria for selecting general-purpose LLMs, making it difficult to determine which model demonstrates superior comprehension of petroleum terminology and engineering reasoning capability. On the other hand, developers of domain-specific models lack quantitative metrics to identify performance deficiencies in particular business scenarios, thereby limiting data-driven optimization and iterative improvement.

Accordingly, the establishment of a domain-specific evaluation benchmark for LLMs in petroleum engineering is essential to overcome the current bottlenecks of model selection uncertainty and unguided optimization. Such a benchmark represents a prerequisite for transitioning LLM applications from laboratory experimentation to large-scale industrial deployment and constitutes the primary motivation of this study.

**1.2 Related Work**

The rapid advancement of Large Language Models (LLMs) has driven the continuous evolution and diversification of evaluation frameworks. Evaluation benchmarks serve not only as essential instruments for measuring model capability but also as critical infrastructure guiding model optimization. In the general-domain context, representative benchmarks such as GLUE, SuperGLUE, HELM, C-Eval, and CMMLU primarily assess natural language understanding, logical reasoning, and cross-disciplinary knowledge question answering, and have played a pivotal role in enhancing general model capabilities [8–12]. Meanwhile, evaluation paradigms have gradually shifted from single-metric accuracy toward multidimensional characterization of model performance, incorporating reasoning capacity, generalization ability, and robustness as key indicators.

Against this backdrop, both academia and industry have recognized that general-purpose benchmarks are insufficient to comprehensively reflect model performance in specialized domains. Consequently, domain-specific evaluation systems have rapidly emerged. In the medical field, datasets such as PubMedQA and MedMCQA focus on medical knowledge comprehension, clinical question answering accuracy, and safety considerations [13–15]. In the legal domain, LegalBench employs a multi-task framework encompassing case reasoning and contract analysis, with emphasis on reasoning consistency and interpretability [16]. In finance, FinEval integrates financial analysis and risk forecasting scenarios, introducing stability and robustness metrics to evaluate reliability in high-risk environments [17]. In engineering and technical domains, benchmarks such as EEE-Bench, PHM-Bench, and ElecBench for the power industry have been proposed, constructing cross-task and multidimensional evaluation frameworks grounded in real-world scenarios including equipment diagnostics and power dispatching, thereby extending benchmark research to complex engineering systems [18,19] In the oil and geoscience domain, FormationEval has focused on geological knowledge question answering by establishing a multiple-choice benchmark targeting petroleum geology, providing an initial exploration of LLM evaluation in specialized geoscience context.

These mature industry-specific benchmarks exhibit several common design principles. First, evaluation tasks are closely aligned with real-world operational scenarios. Second, assessment metrics extend beyond basic accuracy to incorporate multidimensional criteria such as logical consistency and completeness. Third, expert review is commonly combined with large-scale statistical analysis to ensure the reliability and credibility of evaluation outcomes.

In contrast, evaluation frameworks for LLMs in the petroleum industry remain at an early or even nascent stage. Although exploratory applications have been conducted in areas such as exploration and development and production management, existing validation efforts are typically confined to single models or isolated application scenarios and lack systematic design. The absence of a unified and standardized benchmark impedes cross-model and cross-institutional comparisons. Therefore, drawing upon established methodologies from other domains to develop an evaluation benchmark and methodological framework tailored to the specific characteristics of petroleum engineering represents a critical research gap that urgently requires attention.

### 1.3 Contributions of This Study

Current mainstream LLM evaluation frameworks, such as GLUE, SuperGLUE, HELM, C-Eval, and CMMLU, primarily focus on general-purpose capabilities including natural language understanding, logical reasoning, and cross-disciplinary knowledge question-answering. Evaluation metrics typically concentrate on linguistic-level performance measures such as accuracy, precision, recall, and F1 score. While these evaluation systems have matured considerably in assessing general language capabilities of models, their task designs and evaluation objectives cannot be directly mapped to highly specialized industry scenarios such as petroleum engineering.

In contrast, evaluation of domain-specific LLMs emphasizes model adaptability and problem-solving capabilities within specific operational contexts. The petroleum industry possesses distinctive sectoral characteristics, with core tasks often closely tied to engineering decision-making. Requirements for models extend beyond answer correctness to encompass consistency of reasoning logic, conformity of conclusions with engineering practice, and value for application guidance.

Based on the aforementioned background, this study conducts the following work centered on constructing a petroleum engineering LLM evaluation benchmark:

(1) Construction of a petroleum engineering professional evaluation question bank. A systematic review of the core knowledge system and typical operational scenarios in petroleum engineering is conducted to design evaluation questions covering key processes including drilling, production, and reservoir management. A rigorous quality screening mechanism is established to ensure the professionalism and representativeness of the question bank.

(2) Establishment of a multi-dimensional evaluation metric system. Moving beyond the limitations of traditional evaluation that focuses solely on accuracy, multi-dimensional scoring indicators including logical consistency, completeness, and applicability are introduced to comprehensively assess model performance in petroleum professional tasks.

(3) Execution of multi-model comparison and statistical analysis. Multiple mainstream general-purpose LLMs are selected for systematic evaluation. Combined with statistical significance analysis, capability differences among models in the petroleum domain are objectively compared.

The significance of this work is manifested at two levels:

First, it provides a scientific basis for assessing the domain applicability of general-purpose LLMs. Numerous general-purpose LLMs currently exist in the market, and performance differences among models on general-purpose benchmarks do not directly reflect their actual capabilities in the petroleum professional domain. Through the evaluation benchmark constructed in this study, knowledge coverage, reasoning accuracy, and application adaptability of different models in the petroleum domain can be systematically compared, providing quantitative decision-making support for enterprises in introducing and deploying LLMs.

Second, it provides directional guidance for continuous optimization of petroleum industry-specific LLMs. A standardized evaluation system can objectively identify capability boundaries and weak points of models across different professional dimensions, clarify improvement directions, and support iterative enhancement of model capabilities.

Furthermore, establishment of this evaluation system facilitates formation of unified evaluation reference standards at the industry level, promotes systematic comparison across models and institutions, and advances standardized application and robust implementation of artificial intelligence technology in the petroleum engineering domain.

## 2 Scenario Analysis and Evaluation Task Requirements

### 2.1 Analysis of LLM Application Scenarios in the Petroleum Industry

The petroleum engineering domain is characterized by high specialization and interdisciplinary integration, with core operations historically centered on three major areas: drilling engineering, production engineering, and reservoir engineering. Related operational activities involve extensive professional knowledge, engineering experience, and analytical judgment, with information primarily existing in forms such as technical documentation, design specifications, regulatory provisions, and historical case studies. At the current stage, LLM applications in petroleum engineering are predominantly concentrated at the knowledge cognition and analytical support levels, while decision execution for critical processes remains highly dependent on final judgments made by engineering personnel based on professional experience.

In the drilling engineering domain, LLMs are primarily employed to assist in understanding drilling process principles, equipment parameter interpretations, and technical problem analysis under complex operational conditions. For instance, regarding questions related to drilling fluid properties, wellbore stability, or common drilling incident types, models can provide explanatory responses or analytical frameworks based on existing knowledge. Such application scenarios emphasize accurate comprehension of professional terminology and rational articulation of engineering logic by the model.

In production engineering, questions typically involve selection of artificial lift methods, operational mechanisms of production equipment, and relational analysis among production parameters. LLM applications in this domain are primarily manifested in conceptual understanding of production process workflows, analysis of typical problem causes, and comparative descriptions of applicability conditions for different technical measures. The emphasis of model outputs lies in the logical coherence of analytical processes and professional reasonableness of conclusions.

In the reservoir engineering domain, questions generally possess strong theoretical and comprehensive characteristics, such as displacement mechanisms, factors influencing reservoir properties, and development strategy selection. In such scenarios, LLMs more often assume roles of knowledge integration and conceptual reasoning, facilitating understanding of complex problems through articulation of fundamental reservoir engineering principles and typical patterns.

**2.2 Classification of Industry Task Types and Evaluation Requirements**

Unlike general-purpose domains, LLM outputs in industry scenarios often possess decision amplification effects: once a model generates inaccurate or inconsistent conclusions, it may trigger engineering risks, economic losses, or even safety incidents. Therefore, evaluation of industry-specific LLMs must be grounded in real operational tasks and potential risks.

Based on the characteristics of actual problems across the three major professional areas of petroleum engineering, tasks related to industry-specific LLMs can be categorized into the following four core capability types:

(1) Knowledge Comprehension and Conceptual Discrimination

This primarily examines LLMs' capabilities in identifying, interpreting, recalling, and distinguishing fundamental concepts, professional terminology, basic principles, process workflows, and regulatory provisions in petroleum engineering. This constitutes the foundation for constructing higher-level cognitive capabilities. For instance, explanations of drilling fluid performance parameters, fundamental principles of production equipment, or core concepts in reservoir engineering all fall within this category. Term definition questions and portions of multiple-choice and true/false questions primarily correspond to this task type, forming the fundamental component of industry-specific LLM evaluation.

(2) Engineering Problem Analysis and Reasoning

This focuses on LLMs' capabilities in conducting logical analysis, causal diagnosis, trend assessment, risk identification, and inference of complex relationships among multiple factors under given engineering contexts, phenomena, or problem descriptions. In drilling engineering, this manifests as analysis of wellbore stability problem causes; in production engineering, it is reflected in judging causes of production anomalies or applicability of technical measures; in reservoir engineering, it involves analysis of displacement effectiveness differences or development strategy selection logic. Short-answer questions and portions of scenario-based multiple-choice questions primarily cover this task type, constituting an important component for distinguishing capability levels among different models.

(3) Solution Application and Assessment

This examines models' holistic cognition of various engineering solutions, technical measures, or process approaches, with the ability to articulate their working principles, characteristics, applicability conditions, advantages and disadvantages, and potential impacts. It emphasizes models' capabilities in selecting, comparing, and preliminarily evaluating solutions under specific conditions, typically not requiring complex numerical calculations or detailed designs, but rather focusing on qualitative understanding and trade-off analysis of solution merits. Short-answer questions primarily correspond to this task type, with outputs serving as reference bases for engineering personnel's cognitive judgments.

(4) Computation and Quantitative Application

This aims to examine models' capabilities in applying petroleum engineering-related formulas, models, and data for numerical computation, parameter estimation, performance prediction, and quantitative analysis. This includes accurate calculation of physical quantities, engineering parameters, economic indicators, and reasonable interpretation of calculation results. For example, calculating effective reservoir pore volume based on given rock parameters, computing fluid pressure from downhole data, or predicting production changes based on production curves. Multiple-choice questions involving computation and data analysis primarily cover this task type, reflecting models' critical capabilities in quantitative decision-making and refined management in engineering practice.

Different task types exhibit distinct differences in risk characteristics and evaluation focus areas. Knowledge comprehension and conceptual discrimination tasks possess relatively lower risk levels, with model outputs typically serving as references for engineering personnel to acquire and verify industry knowledge. However, deviations in terminology definitions, conceptual boundaries, or principle articulations may still amplify errors in subsequent analysis and reasoning processes. Evaluation of this task type should focus on accuracy of model responses, terminology standardization, and consistency of outputs under different expression conditions. Engineering problem analysis and reasoning tasks require models to conduct causal analysis or logical inference based on understanding of professional context, with results potentially serving as judgment bases for engineering personnel. Risks of this task type primarily manifest in the reliability of model analytical processes. Even if final conclusions appear superficially reasonable, logical leaps, inconsistencies, or omission of critical engineering constraints in reasoning chains may still mislead engineering cognition. Therefore, evaluation processes must attend not only to reasonableness of conclusions but also emphasize logical consistency of model analytical processes, reasoning coherence, and the degree of alignment between conclusions and problem conditions. Solution application and assessment tasks possess risk levels intermediate between the former two, with evaluation focus placed on completeness of response content, professional reasonableness, and clarity of expression structure, while also attending to whether models exhibit oversimplification of complex engineering problems or one-sided emphasis on single factors. Computation and quantitative application tasks may possess higher risks, as quantitative results directly influence decision-making. Evaluation should focus on accuracy of model calculation results, adherence to correct formulas and computational procedures, reasonable interpretation of calculation results and dimensional consistency. Additionally, models' capabilities in handling input data outliers and identifying computational errors are also evaluation priorities.

Evaluation of industry-specific LLMs cannot adopt unified, single-criterion standards. Through comprehensive analysis of task types and risk differences, construction of a multi-dimensional evaluation metric system can provide requirement-driven foundations for evaluation framework design, and constitutes an important prerequisite for ensuring evaluation results genuinely reflect LLM capability performance in petroleum engineering professional scenarios.

# 3 Construction of Petroleum Industry LLM Evaluation Benchmark and Assessment Methodology

To establish a unified, systematic, and practically applicable evaluation benchmark for large language models in the petroleum industry, this study proposes a comprehensive methodological framework encompassing question collection and preprocessing, quality control and filtering, iterative dataset optimization, and multi-model evaluation and validation, as illustrated in Fig. 1. This framework adheres to the standardization of mainstream evaluation benchmarks in the natural language processing community while fully integrating the practical requirements of the three major professional areas in petroleum engineering—drilling, production, and reservoir—aiming to ensure that the evaluation system genuinely reflects the applicability and capability boundaries of LLMs in industry scenarios.

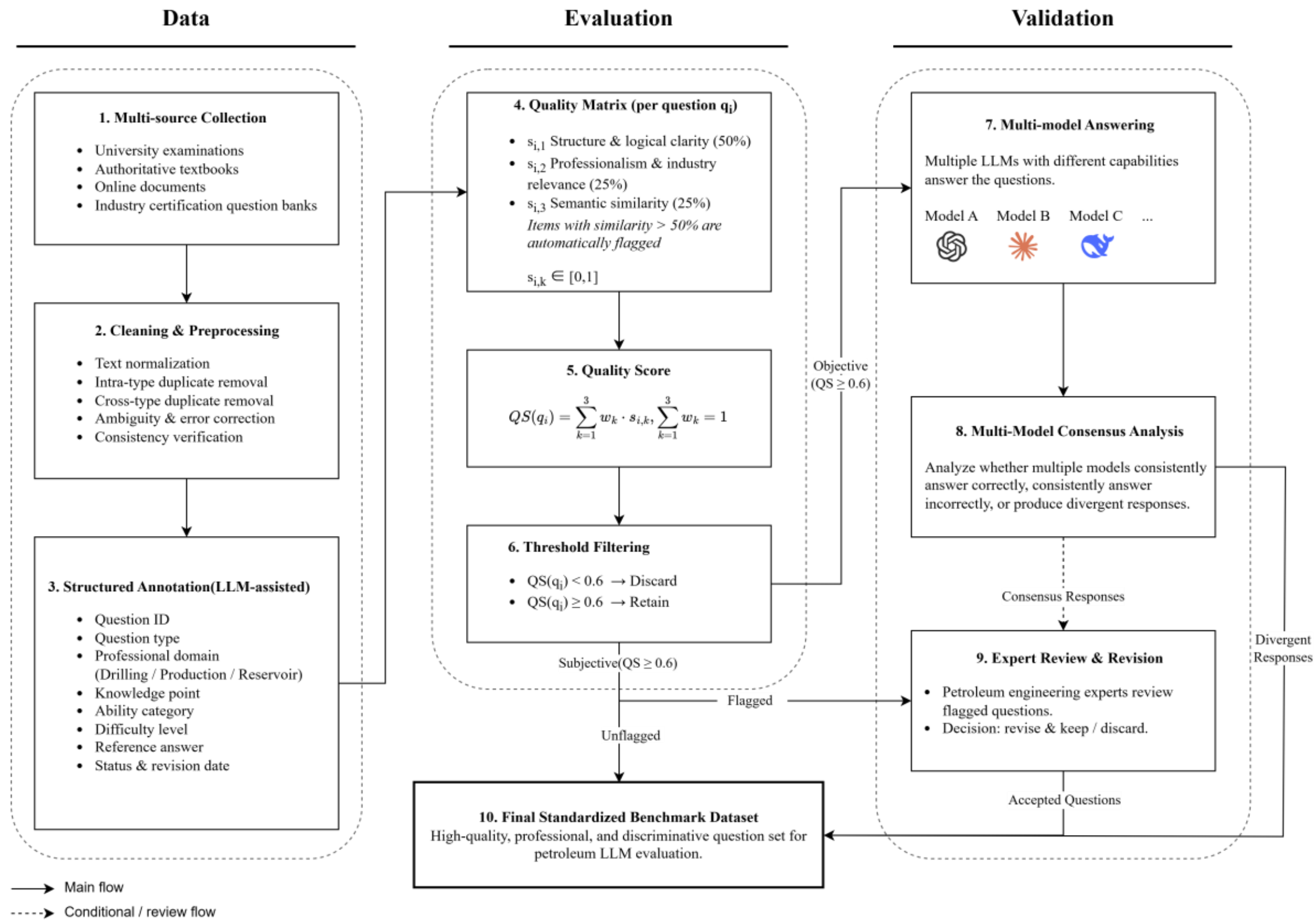

**Fig. 1.** Overall framework of petroleum LLM evaluation benchmark construction and validation

### 3.1 Construction of Evaluation Dataset

The petroleum industry knowledge system is extensive with high professional barriers, encompassing multiple core processes in drilling, production, and reservoir engineering. To ensure coverage and representativeness of the evaluation question bank, this study implemented a systematic data collection and preprocessing workflow.

This study conducted systematic collection from university final examination questions, exercise questions from authoritative textbooks, online documents, and industry qualification examination question banks, constructing an initial corpus pool containing multi-source heterogeneous data. Question type design encompasses four categories to comprehensively test model capabilities: multiple-choice and true/false questions primarily examine models' memory and discrimination abilities regarding fundamental concepts, process principles, and regulatory provisions, corresponding to knowledge comprehension tasks; term definition questions examine accuracy and standardization of models' professional terminology definitions; short-answer questions examine models' logical reasoning, causal analysis, and solution articulation capabilities under specific engineering contexts, corresponding to problem analysis and solution support tasks.

To ensure data consistency and usability, a rigorous cleaning workflow was applied to raw questions. Given that data were primarily organized in structured tabular form, this study adopted a combined strategy of "rule-based filtering + manual verification" for processing:

**Table 1** Question Bank Cleaning Workflow

| Processing Stage | Processing Content | Processing Method |
| --- | --- | --- |
| Text Normalization | Unify text format, remove redundant symbols, garbled characters, repeated spaces and other non-standard characters; standardize mixed Chinese-English text and unit | Batch table processing + manual proofreading |
| | symbols (e.g., m, MPa, °C) | |
| Intra-type Duplication Check | Identify duplicate or highly similar questions within the same question type, merge or eliminate questions with identical examination content | Table filtering and sorting + manual comparison |
| Cross-type Duplication Check | Identify instances where the same definition or knowledge point appears repeatedly across different question types, make selections based on question quality and examination emphasis | Automated cross-comparison + manual comparison |
| Ambiguity and Error Correction | Revise questions with ambiguity, incomplete descriptions, or unreasonable options; supplement incomplete reference answers | Manual review and revision |
| Consistency Verification | Verify correspondence between question stems and reference answers, identify missing answers or labeling errors | Automated identification + multiple rounds of manual proofreading |

Upon completion of the above cleaning workflow, structured label information was established for each question to facilitate subsequent dimensional statistical analysis and stratified sampling. Specific labeling dimensions and methods are shown in the following table:

**Table 2** Structured Label Information for Questions

| Labeling Dimension | Content/Description | Method/Notes |
|---|---|---|
| Question Unique ID | Unique identifier for each question, used for tracking and referencing | Automatically generated |
| Question Text | Complete content of the question | Text |
| Question Type | Multiple-choice/True/false/Short-answer/Term definition | Manual labeling |
| Professional Category | Drilling engineering/Production engineering/Reservoir engineering | Manual labeling |
| Knowledge Point | Specific knowledge points or more detailed professional directions covered by the question | Automated labeling |
| Capability Type Examined | Solution application and assessment/Engineering problem analysis and reasoning/Computation and quantitative application/Knowledge comprehension and conceptual discrimination | Automated labeling with manual sampling verification |
| Difficulty Level | Basic/Intermediate/Advanced | Automated labeling with manual sampling verification |
| Reference Answer | Standard answer to the question | Manual consistency verification |
| Status/Review Marker | Current status of the question in cleaning and review workflow | Internal management field (e.g., pending cleaning, cleaned, pending review, reviewed, disputed, revised, deprecated) |
| Last Revision Date | Date when question text, answer, or metadata was last modified | Manual recording |

Among these, difficulty level was completed using automated LLM labeling. Through designing standardized difficulty assessment prompts, questions were input into LLMs for difficulty determination. Models output corresponding difficulty levels based on factors such as knowledge depth, conceptual complexity, and reasoning requirements involved in questions. Through manual sampling verification, automated labeling results demonstrated high consistency with expert judgments, satisfying the requirements for stratified management of the question bank. The above labeling information was recorded in tabular field format, forming structured question bank data.

Ultimately, all questions and their metadata after cleaning and labeling were uniformly organized and stored in structured tabular form. Each record contains complete question information and metadata labels, providing a standardized data foundation for subsequent quality screening, model evaluation, and result analysis.

**3.2 Question Quality Control and Screening**

To ensure the authority and rigor of the final benchmark, a question quality analysis matrix was established. Candidate items were quantitatively characterized across multiple dimensions, and a threshold-gating mechanism combined with anomaly detection was implemented to enable automatic rejection, retention, and flagging. This strategy minimized manual intervention while preserving only high-quality questions with engineering relevance and evaluation validity. For each candidate question $q_i$, a quality vector $\mathbf{s}_i = [s_{i,1}, s_{i,2}, s_{i,3}]$ was constructed, representing three core dimensions. Each component was normalized to the range $[0, 1]$:

(1) Structural Integrity and Logical Clarity ($s_{i,1}$, weight = 50%)

Rule-based validation and template consistency checks were performed to verify whether the question stem contained necessary background information, whether engineering constraints were explicitly specified, and whether the answer field was complete. For objective questions, the number of options, formatting consistency, and presence of explicit logical conflicts were further examined. Items lacking essential structural components or critical constraints were directly classified as invalid samples.

(2) Domain Specificity and Industry Relevance ($s_{i,2}$, weight = 25%)

Automated classification was conducted to determine whether the question belonged to core petroleum engineering domains (e.g., drilling processes, production equipment, reservoir engineering). Questions exhibiting only general energy knowledge without explicit engineering context were penalized to ensure industry specificity of the benchmark.

(3) Semantic Redundancy Detection ($s_{i,3}$, weight = 25%)

Following field-level deduplication, semantic similarity among questions was computed. Items with similarity exceeding 50% were automatically clustered and flagged for further review.

Based on the quality matrix, a weighted composite Quality Score (QS) was calculated using the weight vector w:

$$QS(q_i) = \sum_{k=1}^{3} w_k \cdot s_{i,k}, \sum_{k=1}^{3} w_k = 1 \quad (1)$$

A question-type-aware threshold-gating strategy was then applied for automatic screening:

If $QS(q_i) < 0.6$: The item is directly excluded from the question bank due to insufficient overall quality.

If $QS(q_i) \geq 0.6$: The item proceeds to differentiated processing based on question type. For subjective questions, the item is first admitted into the candidate pool and then further examined based on its semantic redundancy flag. Items not flagged for redundancy ($s_{i,3} \leq 50\%$) are directly entered into the final question bank, while those flagged ($s_{i,3} > 50\%$) are submitted to the expert panel for review. For objective questions (multiple-choice, true/false), the item proceeds to the multi-model validation stage for further verification.

For objective questions passing the initial quality threshold, a multi-model response validation mechanism was introduced to provide an additional layer of quality assurance. In this validation process, if multiple models consistently produce incorrect responses to the same item, it is flagged for manual verification, with particular attention to potential errors in the reference answer or ambiguity in the question formulation. When models of varying capability levels exhibit consistent deviations on the same item, the issue is more likely attributable to flaws in the question itself rather than model limitations. Items with no anomalies detected are directly entered into the final question bank, while those with detected anomalies are submitted to the expert panel for review. This reverse-validation strategy enables identification of overlooked problematic items that may have passed the initial quality screening, thereby further enhancing the reliability of the benchmark (Fig. 2).

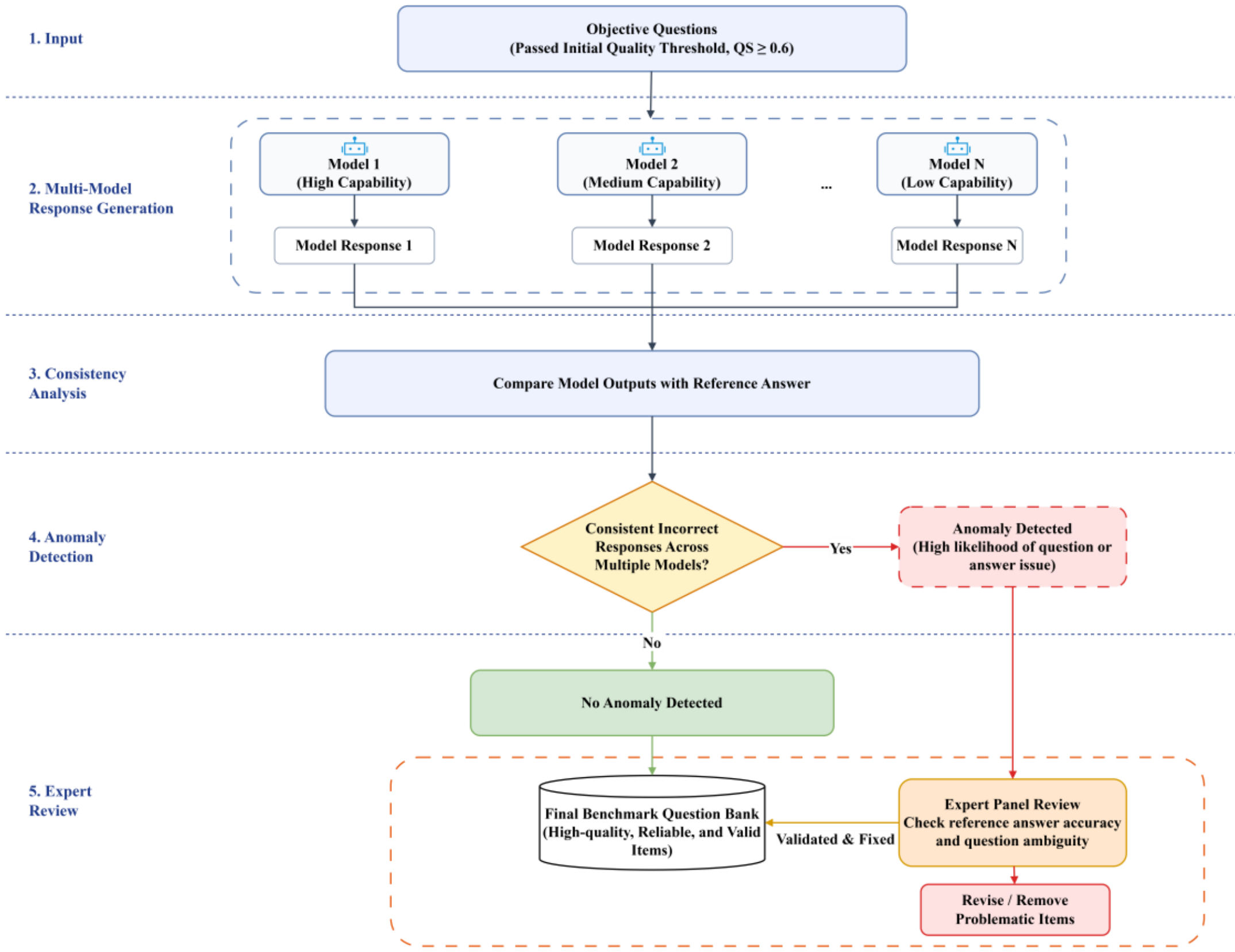

**Fig. 2.** Multi-Model Response Validation Mechanism

The expert review process handles items from two sources: subjective questions flagged for high semantic redundancy ($s_{i,3} > 50\%$) and objective questions exhibiting anomalies in multi-model validation. A panel of petroleum engineering specialists evaluates each flagged item to determine whether revision and inclusion, further review, or permanent removal is appropriate. This procedure ensures a transition from basic data compliance to quality optimization, guaranteeing high standards of scientific rigor, professional relevance, and usability.

Through the sequential processes of data preprocessing, quality-based screening with differentiated handling of subjective and objective questions, multi-model validation for objective items, and expert review, a standardized evaluation question bank was ultimately established, characterized by high rigor, strong domain specificity, and robust discriminative capability.

### 3.3 Results and Structural Analysis of the Question Bank

To further demonstrate the structural rationality and representativeness of the constructed question bank, statistical analysis was conducted from three perspectives: difficulty distribution, knowledge coverage, and capability type distribution.

In terms of difficulty, the question bank exhibits a hierarchical structure dominated by basic and intermediate items, supplemented by a moderate proportion of advanced items. Fig. 3 presents the distribution of the four question types across three difficulty levels (basic, intermediate, and advanced).

Multiple-choice and true–false questions are primarily concentrated at the basic and intermediate levels. Only two advanced multiple-choice questions and six advanced true–false questions were included, indicating that these objective question types mainly serve to assess foundational knowledge recognition, conceptual discrimination, and general logical judgment.

In contrast, short-answer and term-definition questions exhibit greater differentiation in difficulty structure. Among term-definition items, 182 are classified as intermediate and 57 as advanced, reflecting higher requirements for accurate conceptual understanding and standardized technical expression. For short-answer questions, 100 items are intermediate and 43 are advanced, while a substantial proportion of basic-level questions remains, focusing on explanation of fundamental principles and conceptual clarification within open-ended tasks.

Overall, this difficulty distribution ensures stable measurement of foundational competencies while preserving sufficient differentiation for intermediate and advanced capabilities. It thus provides a robust basis for hierarchical analysis of model performance.

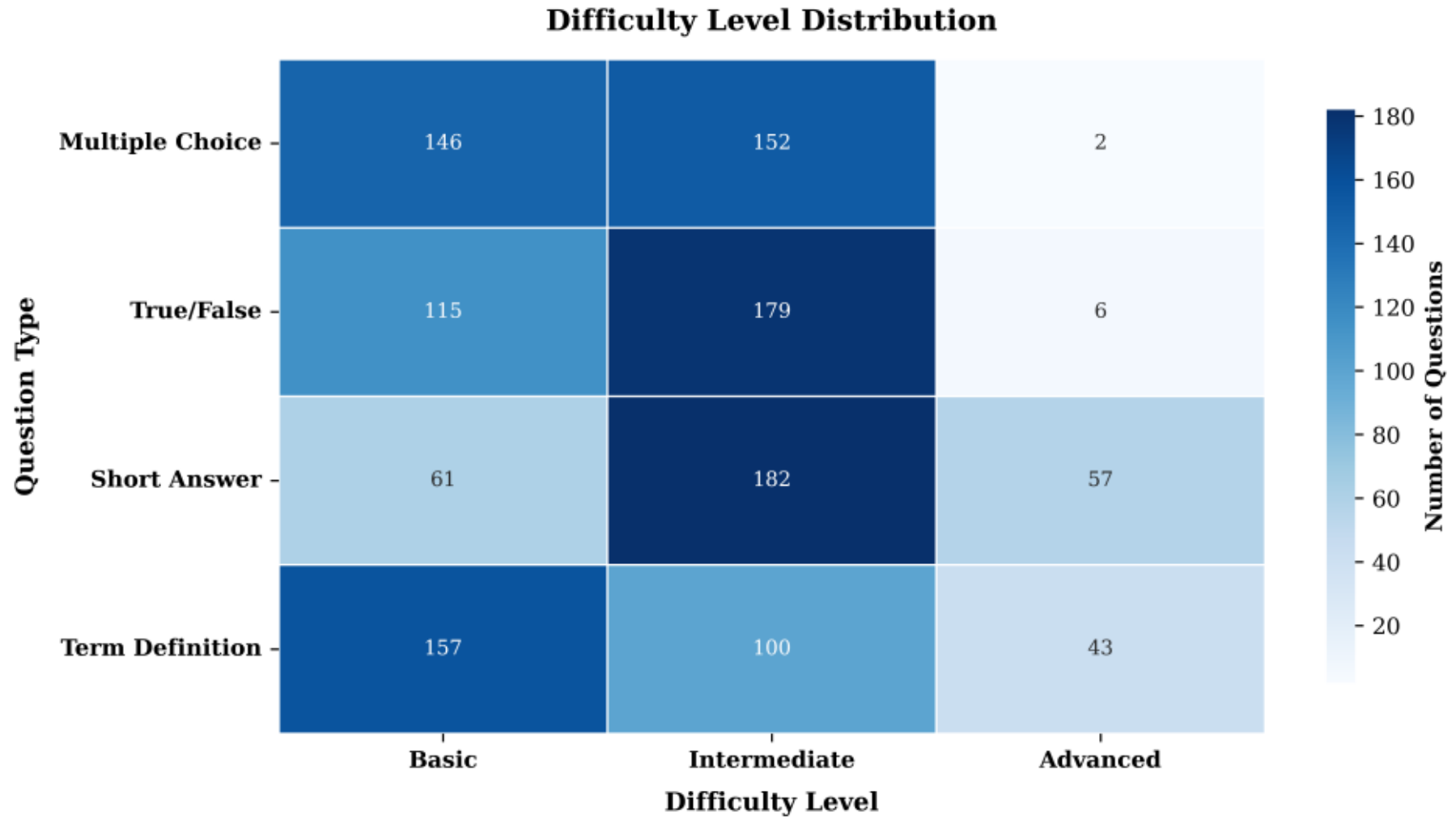

**Fig. 3.** Heatmap of Difficulty Distribution Across Question Types

Beyond question-type structure, an industry-oriented evaluation benchmark must ensure representative knowledge distribution across the three major domains: drilling engineering, production engineering, and reservoir engineering. Fig. 4 illustrates the distribution of the top 12 core knowledge points in each discipline, providing a discipline-level assessment of coverage.

Distinct knowledge concentration patterns are observed across disciplines.

Drilling engineering questions are predominantly concentrated in blowout preventers and control systems, directional drilling, well shut-in and well control operations, well trajectory control, formation pressure and in-situ stress, drill bits, and bottom-hole assembly components. These topics emphasize well control safety, trajectory management, and downhole tool applications.

Production engineering questions frequently involve mechanical artificial lift systems, water injection development, production logging, hydraulic fracturing and acidizing, oil and gas gathering and transportation, multiphase flow in wellbores, and electric submersible pump (ESP) production. The distribution demonstrates strong coverage of artificial lift technologies, stimulation and injection enhancement measures, and surface gathering systems.

Reservoir engineering questions are concentrated in well testing analysis, seepage theory, reservoir fluid characterization, remaining oil evaluation, material balance analysis, capillary pressure, enhanced oil recovery (EOR), and waterflood characteristic curves. This indicates substantial representation in reservoir evaluation, dynamic performance analysis, and displacement mechanisms.

Overall, the distribution of core knowledge points reflects inherent disciplinary distinctions within petroleum engineering while maintaining balanced coverage across the three principal domains.

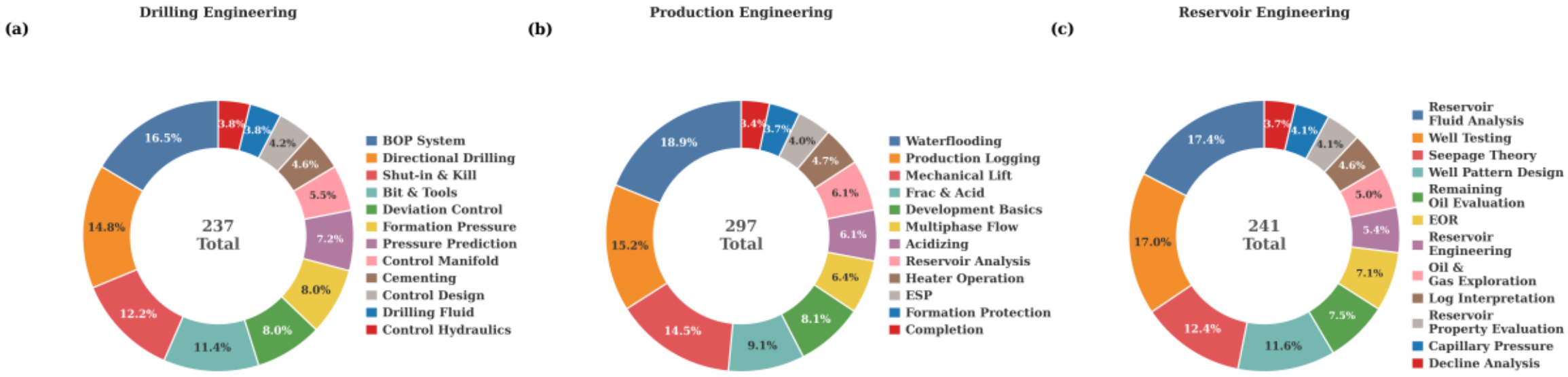

**Fig. 4.** Distribution of Core Knowledge Points in the Three Major Disciplines (Top 12)

The capability structure of the question bank constitutes a critical factor in ensuring the validity of the evaluation benchmark. Fig. 5 illustrates the distribution of four question types across four cognitive capability dimensions: knowledge comprehension and conceptual discrimination, engineering problem analysis and reasoning, scheme application and evaluation, and calculation and quantitative application. Overall, the current question bank is predominantly oriented toward knowledge comprehension and conceptual discrimination; however, significant variation exists among different question types. Multiple-choice and true–false questions exhibit similar capability structures, both dominated by conceptual discrimination tasks, while also incorporating a proportion of engineering problem analysis and reasoning items. In contrast, short-answer questions demonstrate a more balanced capability distribution. Specifically, they comprise 136 items related to knowledge comprehension and conceptual discrimination, 81 items addressing engineering problem analysis and reasoning, 64 items focused on scheme application and evaluation, and 19 items targeting calculation and quantitative application, thereby providing the most comprehensive coverage among all question types. Term-definition questions, by comparison, display a highly concentrated capability distribution. Existing items of this type primarily focus on accurate articulation of professional concepts, terminology definitions, and fundamental principles, which is consistent with their intended assessment function. The calculation and quantitative application dimension remains relatively weak across all question types, suggesting a potential area for future expansion.

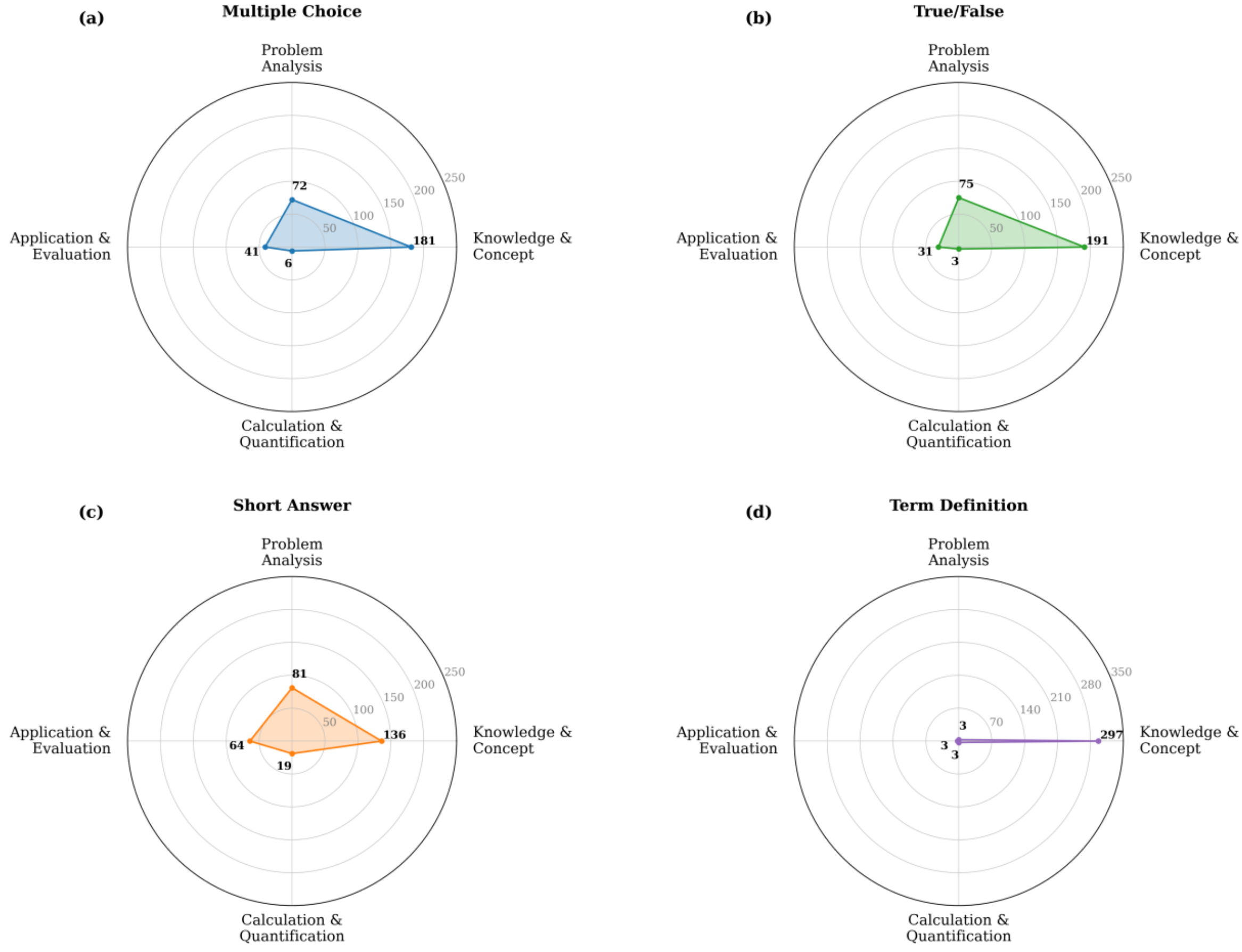

**Fig. 5.** Radar Chart of Capability Distribution Across Question Types

The petroleum industry evaluation question bank developed in this study demonstrates systematic design in terms of difficulty stratification, disciplinary coverage, and capability structure. With respect to difficulty, the dataset is structured with a predominance of basic and intermediate items, supplemented by an appropriate proportion of advanced questions. Regarding disciplinary coverage, the three major domains—drilling engineering, production engineering, and reservoir engineering—are represented by clearly defined and industry-consistent core knowledge points. In terms of capability structure, the question bank effectively spans multiple cognitive levels, ranging from knowledge comprehension to engineering analysis, thereby satisfying diversified evaluation requirements.

### 3.4 Evaluation Experimental Setup and Assessment Metrics

Following establishment of the question bank and screening mechanism, eight representative LLMs were selected, encompassing closed-source commercial models and open-source high-performance models, including GPT, Claude, Gemini, Grok, DeepSeek, Qwen, Doubao, and Kimi. To ensure comparability and fairness of evaluation results, all models were operated under fixed versions with identical invocation strategies to minimize the impact of randomness and implementation differences on results. For objective questions, a standardized zero-shot prompt template was uniformly adopted without additional examples or guiding information, thereby avoiding interference from prompt engineering on model performance. Specific version information for each model is shown in the following table:

**Table 3** Version Information of Evaluated LLMs

| No. | Model | Provider | Version |
|---|---|---|---|
| 1 | GPT | OpenAI | GPT-5.4 |
| 2 | Grok | xAI | Grok-4.2 |

| 3 | Gemini | Google | Gemini-3-Pro |
|---|---|---|---|
| 4 | Claude | Anthropic | Claude-Opus-4.6-Thinking |
| 5 | DeepSeek | DeepSeek | DeepSeek-V3.2-Thinking |
| 6 | Qwen | Alibaba Cloud | Qwen3-Max |
| 7 | Doubao | ByteDance | Doubao-Seed |
| 8 | Kimi | Moonshot AI | Kimi-K2.5 |

Addressing the professional threshold of petroleum engineering, an "LLM-as-a-Judge" automated validation approach was introduced. Currently industry-recognized high-performance models Claude-Sonnet-4.6 and GPT-5.2 were selected as scoring agent models. Both models independently performed complete scoring of 9 candidate responses (including reference answers) for each question. For each response, the arithmetic mean of Claude and GPT scoring results was taken as the final score. If the total score difference between the two models for the same response exceeded 1.0 point, manual expert review was automatically triggered to ensure result reliability. This mechanism effectively circumvents preference biases or comprehension blind spots that may exist in single scoring models, ensuring credibility of evaluation results.

In the evaluation instruction design component, validation models must not only review final answers but also retrieve and compare the <thought> (chain of thought) component of tested models. If a model's logical derivation is erroneous but the conclusion is correct, it will be judged as logically inconsistent and penalized with score reduction. Simultaneously, the evaluation system introduced a core hallucination veto mechanism ($P_{error}$): if a response violates physical laws or exhibits critical dimensional errors, the accuracy score for that question is set to 0, with the total score capped at 2 points.

The weighting system was determined through the Delphi Method: eight petroleum engineering experts were invited to conduct two rounds of anonymous scoring, with consensus weights ultimately adopted. Short-answer question scoring emphasizes logical and analytical capabilities, while term definition scoring emphasizes standardization and completeness of definitions. Multi-dimensional weighted scoring formulas based on semantic understanding were respectively employed:

$$S_{\text{Short-answer}} = 0.40 \times \text{Accuracy} + 0.25 \times \text{Conciseness} + 0.15 \times \text{Logical coherence} + 0.15 \times \text{Relevance}$$

$$\begin{aligned} S_{\text{Term definition}} = {} & 0.25 \times \text{Accuracy} + 0.20 \times \text{Completeness} + 0.15 \times \text{Conciseness} \\ & + 0.10 \times \text{Logical coherence} + 0.10 \times \text{Applicability} + 0.10 \times \text{Professional depth} \\ & + 0.10 \times \text{Currency} \end{aligned}$$

Detailed scoring rubrics for each dimension are presented in the following tables (0–10 point scale):

**Table 4** Short-answer Question Scoring Dimensions

| Dimension | Weight | Scoring Criteria (0–10 Scale) |
|---|---|---|
| Accuracy | 0.40 | 10: Fully accurate and consistent with domain knowledge; 7–9: Minor omissions; 4–6: Noticeable inaccuracies; 1–3: Major errors; 0: Completely incorrect or irrelevant. |
| Conciseness | 0.25 | 10: Clear and direct; 7–9: Slight verbosity; 4–6: Excessively detailed; 1–3: Redundant and unfocused; 0: Extremely verbose or irrelevant. |
| Logicality and Coherence | 0.15 | 10: Clear structure and rigorous logic; 7–9: Minor discontinuities; 4–6: Moderate structural issues; 1–3: Disorganized reasoning; 0: No logical structure. |
| Relevance | 0.15 | 10: Fully aligned with the question; 7–9: Minor deviation; 4–6: Partial relevance; 1–3: Mostly off-topic; 0: Completely irrelevant. |

**Table 5** Scoring Dimensions for Term-Definition Question

| Dimension | Weight | Scoring Criteria (0–10 Scale) |
|---|---|---|
| Accuracy | 0.25 | 10: Fully consistent with standard definitions in petroleum/reservoir engineering; 7–9: Minor deviation; 4–6: Noticeable inaccuracies; 1–3: Fundamental misunderstanding; 0: Completely incorrect. |
| Completeness | 0.20 | 10: Covers all key aspects; 7–9: Minor omissions; 4–6: Missing core elements; 1–3: Severely incomplete; 0: No substantive definition. |
| Conciseness and Clarity | 0.15 | 10: Clear and well-structured; 7–9: Slight verbosity; 4–6: Redundant or partially unclear; 1–3: Disorganized; 0: Incomprehensible. |
| Logicality | 0.10 | 10: Rigorous and coherent; 7–9: Minor logical gaps; 4–6: Structural inconsistencies; 1–3: Illogical; 0: No reasoning structure. |
| Applicability | 0.10 | 10: Closely linked to engineering practice; 7–9: Partial practical relevance; 4–6: Abstract only; 1–3: Minimal practical linkage; 0: No practical relevance. |
| Professional Depth | 0.10 | 10: Demonstrates deep technical understanding; 7–9: Moderate depth; 4–6: Introductory level; 1–3: Superficial; 0: Lacks professional content. |
| Timeliness | 0.10 | 10: Reflects latest research and technologies; 7–9: Some recent developments; 4–6: Traditional perspective; 1–3: Outdated; 0: Obsolete. |

Through the integrated framework of dual-model validation and Delphi-based expert weighting, a comprehensive, logically rigorous, reproducible, and extensible evaluation system for LLMs in the petroleum industry was ultimately established.

# 4 Petroleum Industry LLM Evaluation Results and Analysis

## 4.1 Evaluation Overview and Data Description

Regarding scoring rules, short-answer and term definition questions were scored by evaluation models on a 0–10 point scale according to dimensions; multiple-choice and true/false questions adopted objective scoring methods, tallying the number of correct responses, with 100 questions for each of the three disciplinary domains, totaling a maximum score of 300 points for each. To control evaluation bias, manually written reference answers were simultaneously scored using the same standards, serving as benchmark references. **Table 6** summarizes the four question type scores of the eight models across the complete question bank.

**Table 6** Overall Evaluation Results

| Model | Short Answer (/10) | Term Definition (/10) | Multiple Choice (/300) | True–False (/300) |
|---|---|---|---|---|
| Reference Answer | 6.47 | 6.78 | 300 | 300 |
| Qwen3-Max | 7.76 | 7.84 | 130 | 197 |
| Gemini-3-Pro | 8.01 | 7.83 | 187 | 223 |
| Grok-4.2 | 8.18 | 7.30 | 84 | 190 |
| Claude-Opus-4.6-Thinking | 8.18 | 8.34 | 164 | 211 |
| DeepSeek-V3.2-Thinkin | 7.33 | 8.13 | 134 | 192 |

| g | | | | |
|---|---|---|---|---|
| Doubao-Seed | 7.10 | 6.75 | 144 | 200 |
| Kimi-K2.5 | 7.75 | 8.14 | 196 | 211 |
| GPT-5.4 | 7.74 | 6.74 | 125 | 183 |

Three principal observations can be derived from the overall results:

All models outperformed the reference answers on open-ended question types. This outcome is primarily attributable to the evaluation framework, which incorporated assessment of the models’ reasoning chains (<thought>), whereas the reference answers contained only final conclusions. Consequently, models benefited from additional scoring dimensions related to logical structure and explanatory depth.

Substantial performance variation was observed across question types. No single model ranked first across all four categories, indicating differentiated capability profiles rather than uniform superiority.

Accuracy on objective questions remained relatively low. The highest multiple-choice accuracy was 65.3% (Kimi-K2.5), and the highest true–false accuracy was 74.3% (Gemini-3-Pro). These findings suggest persistent limitations in precise discrimination of factual knowledge within petroleum engineering contexts.

To enable cross-type comparison, scores for each dimension were normalized to a percentage scale and averaged to obtain composite scores, as presented in **Table 7**.

**Table 7** Composite Ranking of Models (Normalized Percentage Mean)

| Rank | Model | Short Answer (%) | Term Definition (%) | Multiple Choice (%) | True–False (%) | Composite (%) |
|---|---|---|---|---|---|---|
| 1 | Gemini-3-Pro | 80.1 | 78.3 | 62.3 | 74.3 | 73.8 |
| 2 | Kimi-K2.5 | 77.5 | 81.4 | 65.3 | 70.3 | 73.6 |
| 3 | Claude-Opus-4.6-Thinking | 81.8 | 83.4 | 54.7 | 70.3 | 72.6 |
| 4 | Qwen3-Max | 77.6 | 78.4 | 43.3 | 65.7 | 66.3 |
| 5 | DeepSeek-V3.2-Thinking | 73.3 | 81.3 | 44.7 | 64.0 | 65.8 |
| 6 | Doubao-Seed | 71.0 | 67.5 | 48.0 | 66.7 | 63.3 |
| 7 | GPT-5.4 | 77.4 | 67.4 | 41.7 | 61.0 | 61.9 |
| 8 | Grok-4.2 | 81.8 | 73.0 | 28.0 | 63.3 | 61.5 |

The composite ranking exhibits a distinct three-tier structure. Tier 1 (72–74%) consists of Gemini-3-Pro, Kimi-K2.5, and Claude-Opus-4.6-Thinking. Although each model demonstrates different strengths, none exhibits a pronounced weakness. Tier 2 (65–67%) includes Qwen3-Max and DeepSeek-V3.2-Thinking. Tier 3 (61–64%) comprises Doubao-Seed, GPT-5.4, and Grok-4.2.

Notably, although Grok-4.2 ranks last overall, it achieved a tied highest score in short-answer questions. Its lower composite ranking is primarily attributable to an extremely low multiple-choice accuracy (28.0%), which substantially reduced its overall score. These results underscore the importance of multidimensional evaluation, as reliance on a single task type would lead to incomplete or potentially misleading conclusions regarding model capability.

**4.2 Analysis by Question Type**

Short-answer questions examine models' capabilities in systematic articulation of petroleum engineering principles, requiring integration of physical mechanisms, engineering logic, and practical relevance. **Table 8** presents the distribution of short-answer question scores for each model across three disciplinary domains. As shown in the table:

(1) Grok-4.2 and Claude-Opus-4.6-Thinking tied for the lead. Although both achieved identical total scores (8.18 points), their areas of strength differ: Grok-4.2 demonstrated the most outstanding performance in production engineering (8.52 points); Claude-Opus-4.6-Thinking excelled in reservoir engineering (8.05 points), exhibiting strong theoretical analysis and mathematical derivation capabilities. Both tied in drilling engineering (both 8.17 points).

(2) Score distribution exhibited disciplinary gradients. Disciplinary mean scores of the eight models demonstrated a consistent trend: production engineering (mean 7.99 points) > drilling engineering (mean 7.74 points) > reservoir engineering (mean 7.53 points). This indicates that reservoir engineering short-answer questions pose the greatest challenge to models, attributable to this domain's involvement of extensive quantitative derivation, seepage mechanics, and multiphase flow theory, imposing higher requirements on models' mathematical processing and formula application capabilities.

(3) Doubao-Seed-1.8-Thinking demonstrated relatively weak performance. Across all three disciplinary domains, Doubao-Seed-1.8-Thinking's short-answer question scores ranked last or second-to-last, with an overall mean score of only 7.10 points, differing from the first-place score by over 1 point, primarily manifesting as insufficient response depth and missing critical engineering details.

**Table 8** Short-answer Question Scores by Discipline

| Model | Production Eng. | Reservoir Eng. | Drilling Eng. | Overall Mean | Rank |
|---|---|---|---|---|---|
| Grok−4. 2 | **8. 52** | 7. 86 | **8. 17** | **8. 18** | 1 |
| Claude-Opus-4.6-Thinking | 8. 33 | **8. 05** | **8. 17** | **8. 18** | 1 |
| Gemini−3−Pro | 8. 16 | 7. 77 | 8. 08 | 8. 01 | 3 |
| Qwen3−Max | 8. 12 | 7. 53 | 7. 65 | 7. 76 | 4 |
| Kimi−K2. 5 | 7. 80 | 7. 63 | 7. 80 | 7. 75 | 5 |
| GPT−5. 4 | 7. 96 | 7. 50 | 7. 76 | 7. 74 | 6 |
| DeepSeek-V3.2-Thinking | 7. 57 | 7. 11 | 7. 32 | 7. 33 | 7 |
| Doubao−Seed | 7. 48 | 6. 86 | 6. 95 | 7. 10 | 8 |

Term definition questions examine models' definitional accuracy and conceptual completeness regarding petroleum engineering professional terminology. As shown in **Table 9**:

(1) Claude-Opus-4.6-Thinking led with a significant advantage. Its total score of 8.34 points exceeded second-place Kimi-K2.5 by nearly 0.2 points, and achieved the highest scores in both reservoir engineering (8.50 points) and drilling engineering (8.42 points). Analysis of its response characteristics revealed that Claude-Opus-4.6-Thinking possesses clear definitional boundaries, complete physical connotations, and reasonable engineering extensions in terminology definitions, capable of embedding concepts within complete engineering contexts.

(2) Grok-4.2 experienced substantial ranking decline. While Grok-4.2 tied for first place in short-answer questions, it ranked only sixth in term definitions (7.30 points). This reveals an important capability differentiation phenomenon: Grok-4.2 excels in systematic principle analysis and argumentation, such as generating long texts and deep reasoning, but exhibits deficiencies in precise definition and concise summarization, with responses often over-expanding and deviating from core requirements of terminology definitions.

(3) GPT-5.4 and Doubao-Seed-1.8-Thinking formed a distinct gap at the bottom. A difference of 0.55 points exists between their scores (6.74 and 6.75 points) and sixth-place Grok-4.2 (7.30 points), with primary issues manifesting as: insufficiently precise terminology definitions, conceptual confusion, or missing critical qualifying conditions. Particularly in drilling engineering, GPT-5.4 achieved only 6.49 points, the lowest score among all tested models for this discipline, reflecting this model's obviously insufficient mastery of the drilling professional terminology system.

**Table 9** Term Definition Question Scores by Discipline

| Model | Production Eng. | Reservoir Eng. | Drilling Eng. | Overall Mean | Rank |
|---|---|---|---|---|---|
| Claude-Opus-4.6-Thinking | 8. 10 | **8. 50** | **8. 42** | **8. 34** | 1 |
| Kimi−K2. 5 | **8. 17** | 8. 23 | 8. 03 | 8. 14 | 2 |
| DeepSeek-V3.2-Thinking | 8. 11 | 8. 06 | 8. 22 | 8. 13 | 3 |
| Qwen3−Max | 7. 81 | 7. 92 | 7. 81 | 7. 84 | 4 |
| Gemini−3−Pro | 7. 83 | 7. 88 | 7. 79 | 7. 83 | 5 |
| Grok−4. 2 | 7. 54 | 7. 38 | 6. 97 | 7. 30 | 6 |
| Doubao−Seed | 6. 93 | 6. 73 | 6. 60 | 6. 75 | 7 |
| GPT−5. 4 | 6. 99 | 6. 73 | 6. 49 | 6. 74 | 8 |

Multiple-choice questions constitute a critical question type for examining models' factual knowledge precision, requiring models to identify fixed correct answers among distractors, with low error tolerance. As shown in **Table 10**:

(1) Model performance stratification was significant. Kimi-K2.5 demonstrated optimal performance in factual discrimination capability testing, ranking first with an accuracy of 196/300 (65.3%), with balanced score distribution across the three disciplinary domains of reservoir engineering, drilling engineering, and production engineering (65-66 questions), exhibiting no obvious disciplinary bias. Gemini-3-Pro ranked second with an accuracy of 187/300 (62.3%), similarly demonstrating good cross-disciplinary stability. Together they constitute the "first tier" of this evaluation, with accuracy presenting significant advantages over other models ($p<0.05$).

(2) Grok-4.2 exhibited significant task-type adaptability differences. Grok-4.2's performance in multiple-choice questions was anomalous, with an overall accuracy of only 28.0% (84/300). Particularly in drilling engineering, its accuracy dropped to 11% (11/100 questions), forming a stark contrast with its tied first-place performance in subjective question-answering. Error case analysis revealed that this model exhibits over-reasoning phenomena during multiple-choice question answering—after conducting deep semantic analysis of distractor options, it deviates from correct answers based on direct knowledge base retrieval. This demonstrates the negative transfer effect that deep reasoning capabilities may produce in specific task scenarios, namely, in contexts requiring precise factual retrieval rather than complex reasoning, the model's long-chain reasoning mechanism becomes an interference factor.

(3) Insufficient drilling engineering knowledge coverage. Drilling engineering discipline exhibited a universal low accuracy phenomenon in this evaluation. Except for Kimi-K2.5 (65 questions), scores of other models in this discipline were significantly lower than in reservoir engineering and production engineering (one-way ANOVA, $F>4.5$, $p<0.01$). Among them, drilling engineering accuracy for Grok-4.2 (11 questions), DeepSeek-V3.2-Thinking (23 questions), and GPT-5.4 (30 questions) were all below 35%. This phenomenon can be attributed to the highly concrete characteristics of drilling engineering knowledge—involving extensive fine-grained engineering knowledge such as specific model drilling tool parameters, operational specification codes, and industry standard provisions—while general-purpose corpora exhibit obviously insufficient coverage density of such professional details, resulting in models lacking effective knowledge support for such questions.

(4) Systematic precision deficiencies are universally present. Accuracy of all tested models did not exceed the 70% threshold, reflecting systematic limitations of current LLMs in professional domain knowledge precision. Qualitative analysis of incorrect answer samples revealed that models universally exhibited characteristics of "sufficient conceptual-level understanding but insufficient fact-level precision"—capable of accurately articulating relevant principles but frequently erring in multiple-choice questions requiring discrimination of approximate concepts or precise parameters. This reveals the mismatch problem between semantic representation density and factual granularity in pre-training corpora, namely, conceptual knowledge acquired by models through large-scale text learning remains insufficient to support reliable discrimination of precise facts in professional domains.

**Table 10** Number of Correct Multiple-choice Questions by Discipline (Maximum 100 questions each)

| Model | Production Eng. | Reservoir Eng. | Drilling Eng. | Total (/300) | Overall Mean | Rank |
|---|---|---|---|---|---|---|
| Kimi−K2. 5 | 66 | **65** | **65** | **196** | **65. 3** | 1 |
| Gemini−3−Pro | **66** | **66** | 56 | 187 | 62. 3 | 2 |
| Claude-Opus-4.6-Thinking | 61 | 61 | 42 | 164 | 54. 7 | 3 |
| Doubao−Seed | 52 | 53 | 39 | 144 | 48. 0 | 4 |
| DeepSeek-V3.2-Thinking | 58 | 53 | 23 | 134 | 44. 7 | 5 |
| Qwen3−Max | 44 | 46 | 40 | 130 | 43. 3 | 6 |
| GPT−5. 4 | 54 | 57 | 30 | 125 | 41. 7 | 7 |
| Grok−4. 2 | 40 | 33 | 11 | 84 | 28. 0 | 8 |

True/false questions require models to make binary judgments on statement correctness, examining knowledge accuracy and capability to discriminate common misconceptions. Overall performance on true/false questions surpassed multiple-choice questions, though the difference magnitude was smaller than expected, indicating that models do not simply rely on probabilistic advantages. As shown in **Table 11**:

(1) Gemini-3-Pro demonstrated optimal performance in true/false questions. Leading with an accuracy of 223/300 (74.3%), it achieved the highest score across all models in drilling engineering with 81/100. Gemini-3-Pro's overall advantage in objective questions is prominent, indicating this model possesses strong factual knowledge reserves and true/false discrimination capabilities.

(2) Claude-Opus-4.6-Thinking and Kimi-K2.5 tied for second place. Both achieved 211/300 (70.3%), but with different areas of strength: Claude-Opus-4.6-Thinking achieved the highest score across all models in reservoir engineering true/false questions with 70/100, while Kimi-K2.5 performed better in production engineering (73 questions) and drilling engineering (76 questions).

(3) Inter-model gaps were smaller than in multiple-choice questions. Only a 13.3 percentage point difference exists between the highest accuracy (74.3%) and lowest accuracy (61.0%) in true/false questions, whereas in multiple-choice questions this gap was 37.3 percentage points (65.3%–28.0%). This indicates that in binary judgment tasks, the degree of model capability differentiation is smaller, with all models possessing basic true/false discrimination literacy, but with greater differences in refined multi-option discrimination capabilities.

(4) GPT-5.4 demonstrated poor performance. In true/false questions, GPT-5.4 ranked last with 183/300 (61.0%), and achieved only 58/100 in production engineering, the lowest single-discipline score across all models. Combined with its second-to-last performance in multiple-choice questions, GPT-5.4's overall performance on petroleum engineering objective questions is at a disadvantage.

**Table 11** Number of Correct True/False Questions by Discipline (Maximum 100 questions each)

| Model | Production Eng. | Reservoir Eng. | Drilling Eng. | Total (/300) | Overall Mean | Rank |
|---|---|---|---|---|---|---|
| Gemini−3−Pro | **75** | **67** | **81** | **223** | **74. 3** | 1 |
| Claude-Opus-4.6-Thinking | 70 | **70** | 71 | 211 | 70. 3 | 2 |
| Kimi−K2. 5 | **73** | 62 | **76** | 211 | 70. 3 | 2 |
| Doubao−Seed | 65 | 66 | 69 | 200 | 66. 7 | 4 |
| Qwen3−Max | 64 | 60 | 73 | 197 | 65. 7 | 5 |
| DeepSeek-V3.2-Thinking | 64 | 60 | 68 | 192 | 64. 0 | 6 |

| Grok-4. 2 | 64 | 60 | 66 | 190 | 63. 3 | 7 |
|---|---|---|---|---|---|---|
| GPT-5. 4 | 58 | 59 | 66 | 183 | 61. 0 | 8 |

**4.3 Analysis of Inter-disciplinary Differences**

Production engineering represents the disciplinary domain with the best overall model performance among the three disciplines. The eight-model mean score for short-answer questions was 7.99 points, and for term definitions 7.69 points, both the highest among the three disciplines. Average accuracy for multiple-choice questions was 52.5%, and for true/false questions 66.5%, similarly leading the other two disciplines. This indicates that production engineering knowledge is more closely aligned with common production practice content in model training corpora, with relatively intuitive concepts and strong descriptive process workflows, reducing the difficulty of model comprehension and response.

Reservoir engineering poses the greatest challenge to models among the disciplinary domains. The mean score for short-answer questions was 7.53 points, the lowest among the three disciplines, directly related to reservoir engineering's involvement of extensive quantitative derivation content such as seepage mechanics equations, relative permeability curve analysis, and material balance calculations. Claude-Opus-4.6-Thinking demonstrated outstanding performance in short-answer questions in this domain. In term definitions, Claude-Opus-4.6-Thinking achieved the highest score of 8.50 points, with definitional precision for reservoir engineering professional terminology reaching quality approaching textbook level. In multiple-choice questions, the model average accuracy for reservoir engineering was 49.25%, essentially comparable to production engineering, but Grok-4.2 achieved only 33/100 in this domain, revealing severe deficiencies in reservoir engineering quantitative knowledge.

Drilling engineering exhibited the most extreme model differentiation in multiple-choice questions. A difference of 54 questions exists between highest-scoring Kimi-K2.5 and lowest-scoring Grok-4.2, with a high divergence rate. The knowledge system of drilling engineering is characterized by highly engineering-practical features: drilling assembly design, drilling fluid formulations, well control operational procedures, casing program design, and other content involving extensive industry standards, empirical parameters, and operational specifications. These knowledge elements appear in general internet corpora at frequencies far lower than theoretical content in production and reservoir engineering. Therefore, drilling multiple-choice questions serve as critical validation indicators for testing model industry knowledge depth. In true/false questions, Gemini-3-Pro achieved the highest score across all models in drilling with 81/100, with Kimi-K2.5 ranking second (76/100), indicating these two models possess relative advantages in breadth coverage of drilling engineering knowledge. In short-answer questions, Grok-4.2 and Claude-Opus-4.6-Thinking tied for first place, again confirming the capability decoupling phenomenon between open-ended analytical questions and closed-ended knowledge questions.

Based on the above analysis, significant and systematic differentiation exists between LLMs' knowledge generation capabilities and knowledge discrimination capabilities. This differentiation may originate from two aspects: first, generative tasks allow models to utilize their language organization capabilities to embellish or expand answers, partially masking knowledge blind spots, whereas multiple-choice questions require precise identification of the sole correct option, with no ambiguous space. Second, some models with strong reasoning capabilities tend to over-analyze options in multiple-choice questions, deviating from intuitively correct answers after repeated deliberation.

**4.4 Comparison Between International and Chinese Models**

From the perspective of model attribution, Chinese models among the evaluated models include Qwen3-Max, DeepSeek-V3.2-Thinking, Doubao-Seed-1.8-Thinking, and Kimi-K2.5, while international models include Gemini-3-Pro, Grok-4.2, Claude-Opus-4.6-Thinking, and GPT-5.4.

**Table 12** Comparison of Average Scores by Question Type Between Chinese and International Models

| Dimension | Chinese Model Mean | International Model Mean | Advantage |
|---|---|---|---|
| Short-answer Questions | 7. 48 | 8. 03 | International models |

| Term Definitions | 7. 72 | 7. 55 | Chinese models |
| --- | --- | --- | --- |
| Multiple-choice Accuracy | 50. 3% | 46. 7% | Chinese models |
| True/false Accuracy | 66. 7% | 67. 2% | Essentially comparable |

On tasks such as short-answer questions requiring in-depth analysis and systematic exposition, international models possess a leading advantage of approximately 0.55 points, potentially related to their larger-scale multilingual training and reasoning optimization. In multiple-choice questions, Chinese models overtook with an advantage of approximately 3.6 percentage points, with Kimi-K2.5 (65.3%) topping the multiple-choice rankings, possibly benefiting from Chinese models' more comprehensive coverage of Chinese petroleum engineering textbooks and literature corpora. Chinese models demonstrate slight advantages in term definitions, with true/false questions essentially comparable.

Overall, no significant systematic gap exists in capability comparison between Chinese and international models in the petroleum engineering domain, with each excelling in different question types and areas. Particularly noteworthy is that Kimi-K2.5 ranked second in comprehensive rankings with only a 0.2 percentage point difference (73.6% vs 73.8%), with its outstanding performance in multiple-choice questions providing empirical support for the competitiveness of Chinese models.

**4.5 Model Response Speed Analysis**

Response speed is also an important metric for measuring the practical application value of LLMs. This study selected two representative question types—multiple-choice and short-answer questions—to conduct response speed analysis. Multiple-choice question answers are highly standardized with nearly constant output token counts, effectively eliminating interference from generated content length on response time. Therefore, response time for multiple-choice questions primarily reflects model knowledge retrieval efficiency, reasoning speed, and API service performance, providing a fair baseline for cross-model speed comparison. Short-answer questions require generation of structured long texts of 200-500 tokens, involving principle exposition, logical derivation, and engineering analysis, more closely approximating complex question-answering scenarios in practical applications. Their response time is comprehensively influenced by reasoning depth, content generation speed, and output length, capable of reflecting end-to-end performance of models under real workloads. Test samples consisted of 300 questions (100 questions for each of the three disciplines).

**Table 13** presents statistical results for average response time, maximum response time, and minimum response time of each model in multiple-choice question tasks.

**Table 13** Response Time Statistics for Models in Multiple-choice Question Tasks

| Model | Average Time (s) | Maximum Time (s) | Minimum Time (s) |
| --- | --- | --- | --- |
| GPT-5. 4 | 3. 297 | 18. 801 | 1. 860 |
| Grok-4. 2 | 4. 355 | 26. 111 | 1. 963 |
| Qwen3-Max | 7. 571 | 70. 852 | 1. 974 |
| DeepSeek-V3.2-Thinking | 9. 434 | 104. 998 | 1. 537 |
| Claude-Opus-4.6-Thinking | 11. 305 | 118. 233 | 2. 110 |
| Doubao-Seed-1. 8-Thinking | 15. 842 | 84. 555 | 3. 094 |
| Gemini-3-Pro | 34. 771 | 122. 181 | 5. 336 |
| Kimi-K2. 5 | 41. 438 | 126. 010 | 1. 662 |

From the multiple-choice question results, efficiency differences among models in short-response scenarios are significant. GPT-5.4 and Grok-4.2 constitute the fastest first tier in response speed, with average response times of 3.297s and 4.355s respectively, notably lower than other models. Qwen3-Max, DeepSeek-V3.2-Thinking, and Claude-Opus-4.6-Thinking form the intermediate tier, with average response times generally distributed in the 7–12s range, demonstrating relatively moderate response efficiency. In contrast, average response times for Gemini-3-Pro and Kimi-K2.5 reached 34.771s and 41.438s respectively, significantly higher than other models, indicating substantial reasoning or generation latency in objective question scenarios.

From the perspective of response time fluctuation range, obvious differences in stability exist among models. Taking GPT-5.4 as an example, its maximum response time is only 18.801s, the lowest level among all models, indicating that its response in multiple-choice question tasks is not only fast but also exhibits small fluctuation. In contrast, maximum response times for DeepSeek-V3.2-Thinking, Claude-Opus-4.6-Thinking, Gemini-3-Pro, and Kimi-K2.5 all exceeded 100s, manifesting relatively obvious long-tail latency phenomena. This indicates that some models may trigger more complex internal reasoning processes or higher system scheduling overhead on individual questions, resulting in significantly prolonged response times.

**Table 14** presents statistical results for response times of each model in short-answer question tasks.

**Table 14** Response Time Statistics for Models in Short-answer Question Tasks

| Model | Average Time (s) | Maximum Time (s) | Minimum Time (s) |
|---|---|---|---|
| Grok−4. 2 | 9. 989 | 86. 658 | 5. 456 |
| Qwen3−Max | 12. 555 | 32. 261 | 8. 036 |
| Kimi−K2. 5 | 20. 755 | 111. 491 | 4. 616 |
| DeepSeek-V3.2-Thinking | 20. 799 | 177. 995 | 7. 114 |
| Claude-Opus-4.6-Thinking | 22. 369 | 162. 291 | 11. 296 |
| Gemini-3-Pro | 24. 288 | 95. 482 | 8. 451 |
| GPT−5. 4 | 25. 539 | 51. 891 | 9. 736 |
| Doubao−Seed−1. 8−Thinking | 26. 855 | 104. 541 | 8. 061 |

Compared to multiple-choice questions, average response times for short-answer questions increased significantly overall, consistent with task characteristics requiring generation of longer texts and organization of more complex argumentative structures. Grok-4.2 demonstrated the highest response efficiency in short-answer question tasks with an average response time of 9.989s; Qwen3-Max ranked second at 12.555s, both notably faster than other models. In contrast, average response times for Doubao-Seed-1.8-Thinking, GPT-5.4, and Gemini-3-Pro all exceeded 24s, indicating relatively high time costs for these models in open-ended question-answering scenarios.

In terms of maximum response time, fluctuations among models are more pronounced in short-answer questions. Maximum response times for DeepSeek-V3.2-Thinking and Claude-Opus-4.6-Thinking reached 177.995s and 162.291s respectively, far exceeding their average response times, exhibiting significant long-tail phenomena. Such fluctuations are typically associated with long-chain reasoning, complex generation paths, and underlying service scheduling. Notably, although Qwen3-Max does not have the lowest average response time, its maximum response time is only 32.261s, demonstrating relatively good stability among all models.

Integrating results from multiple-choice and short-answer questions reveals that no simple positive correlation exists between model response efficiency and answer quality. In multiple-choice question tasks, although Kimi-K2.5 and Gemini-3-Pro achieved first and second place in accuracy respectively, their response times also ranked at the highest levels among all models, indicating that higher factual discrimination capabilities may be accompanied by longer internal reasoning or more complex generation processes. Conversely, GPT-5.4 demonstrated high response efficiency in multiple-choice questions, but its objective question accuracy did not enter the leading tier; although Grok-4.2 similarly responded quickly, its multiple-choice accuracy was notably low, indicating that high response efficiency does not necessarily correspond to high factual discrimination capability.

In short-answer question tasks, Grok-4.2 and Claude-Opus-4.6-Thinking tied for first place in scores, but their performance in response efficiency differed. Grok-4.2 maintained relatively fast average response time while achieving high scores, whereas Claude-Opus-4.6-Thinking's time cost was notably higher. This indicates that even when open-ended question-answering quality is comparable, internal reasoning mechanisms and generation paths may still exhibit significant differences among models.

Overall, a relatively clear trade-off relationship exists between accuracy and response efficiency for LLMs in the petroleum engineering domain. Faster models are more suitable for high-frequency interaction, real-time question-answering, and lightweight deployment scenarios, while models with higher accuracy but slower response are more appropriate for professional evaluation, decision support, and application scenarios with high knowledge precision requirements. Therefore, in practical applications, model selection should involve comprehensive trade-offs considering task type, timeliness requirements, and precision needs, rather than judgments based solely on single evaluation metrics.

It should be noted that although all models were invoked through a unified API platform, response times may still be influenced by external factors such as network transmission conditions, underlying service scheduling strategies, and concurrent loads. Therefore, results in this section are primarily used to reflect relative response efficiency of models under a unified testing environment, and should not be simply regarded as the sole representation of their absolute reasoning speed.

# 5 Conclusions and Prospects

Oriented toward petroleum engineering professional scenarios, this study constructed a professional question bank encompassing three core areas—production engineering, reservoir engineering, and drilling engineering—and accordingly established a multi-dimensional LLM evaluation framework covering both subjective and objective questions. Research results indicate that general-purpose LLMs already possess certain capabilities in knowledge retrieval and text generation in the petroleum engineering domain, but their professional capabilities do not exhibit synchronized improvement across a single dimension; rather, they manifest relatively obvious structural differences. Particularly across different tasks such as open-ended responses, concept explanation, factual discrimination, and quantitative derivation, model advantages are not consistent, indicating that LLM evaluation in petroleum engineering scenarios cannot rely solely on single question types or single indicators, but should adopt more targeted comprehensive evaluation methods.

The primary significance of this study lies in providing a quantifiable and reproducible evaluation benchmark for the petroleum engineering domain. Compared to relatively mature industry evaluation systems in domains such as medicine, law, and finance, petroleum engineering has long lacked systematic evaluation tools for LLMs. The evaluation framework constructed in this study not only provides unified references for horizontal model comparison, but also establishes a methodological foundation for subsequent industry model fine-tuning, capability diagnosis, and application selection. The research also demonstrates that in professional scenarios, language generation quality of models cannot fully represent the reliability of their knowledge judgments—a finding that holds practical implications for training objective setting and deployment implementation of petroleum engineering LLMs.

This study has certain limitations. First, the subjective question scoring process relies primarily on LLM evaluation. Although this method possesses high efficiency, it may still harbor implicit preferences for answer length, structural completeness, and expression fluency, thereby affecting the absolute objectivity of scoring results. Second, although the current question bank covers core petroleum engineering knowledge, incorporation of emerging areas such as unconventional oil and gas development, smart oilfields, and carbon capture and storage remains insufficient, with room for further refinement of question difficulty levels and scenario complexity. Additionally, this study primarily focuses on model output quality and reasoning latency, with key indicators such as computational cost, multi-turn interaction capabilities, and tool invocation capabilities not yet incorporated into a unified framework.

Future research can be further deepened in the following aspects: first, continuously expanding and updating question bank content to enhance the evaluation system's coverage of new technologies, new processes, and composite tasks; second, incorporating indicators such as cost and stability to construct a comprehensive evaluation system more closely aligned with actual deployment requirements; third, extending evaluation scenarios from static question-answering to more complex tasks such as multi-turn interaction, long-document comprehension, real-time retrieval, and external tool invocation, thereby enhancing the explanatory power of test results for real engineering applications. With continuous improvement of evaluation content and methods, this benchmark is expected to develop into an important support tool for LLM research and development, selection, and optimization in the petroleum engineering domain, and to provide foundational references for industry intelligent transformation.

# Acknowledgements

The authors acknowledge fundings from the Jiangsu Provincial Qinglan Project (2024), the National Natural Science Foundation of China (U24B6003) and the Taishan Scholar Program of Shandong Province (tsqn202408088).

## References

[1] Brown T, Mann B, Ryder N, et al. Language models are few-shot learners[J]. Advances in Neural Information Processing Systems, 2020, 33: 1877-1901.

[2] Chowdhery A, Narang S, Devlin J, et al. PaLM: Scaling language modeling with pathways[J]. Journal of Machine Learning Research, 2023, 24(240): 1-113.

[3] Liu P, Yuan W, Fu J, et al. Pre-train, prompt, and predict: A systematic survey of prompting methods in natural language processing[J]. ACM Computing Surveys, 2023, 55(9): 1-35.

[4] Raffel C, Shazeer N, Roberts A, et al. Exploring the limits of transfer learning with a unified text-to-text transformer[J]. Journal of Machine Learning Research, 2020, 21(140): 1-67.

[5] Ouyang L, Wu J, Jiang X, et al. Training language models to follow instructions with human feedback[J]. Advances in Neural Information Processing Systems, 2022, 35: 27730-27744.

[6] Hu E J, Shen Y, Wallis P, et al. LoRA: Low-rank adaptation of large language models[C]//International Conference on Learning Representations. 2022.

[7] Dettmers T, Pagnoni A, Holtzman A, et al. QLoRA: Efficient finetuning of quantized LLMs[J]. Advances in Neural Information Processing Systems, 2023, 36: 10088-10115.

[8] Wang A, Singh A, Michael J, et al. GLUE: A multi-task benchmark and analysis platform for natural language understanding[C]//Proceedings of the 2018 EMNLP Workshop BlackboxNLP: Analyzing and Interpreting Neural Networks for NLP. Brussels, Belgium: Association for Computational Linguistics, 2018: 353-355.

[9] Wang A, Pruksachatkun Y, Nangia N, et al. SuperGLUE: A stickier benchmark for general-purpose language understanding systems[J]. Advances in Neural Information Processing Systems, 2019, 32: 3266-3280.

[10] Liang P, Bommasani R, Lee T, et al. Holistic evaluation of language models[J]. Transactions on Machine Learning Research, 2023.

[11] Huang Y, Bai Y, Zhu Z, et al. C-Eval: A multi-level multi-discipline Chinese evaluation suite for foundation models[J]. Advances in Neural Information Processing Systems, 2023, 36: 62991-63010.

[12] Li H, Zhang Y, Koto F, et al. CMMLU: Measuring massive multitask language understanding in Chinese[C]//Findings of the Association for Computational Linguistics: ACL 2024. Bangkok, Thailand, 2024: 11260-11285.

[13] Jin Q, Dhingra B, Liu Z, et al. PubMedQA: A dataset for biomedical research question answering[C]//Proceedings of the 2019 Conference on Empirical Methods in Natural Language Processing and the 9th International Joint Conference on Natural Language Processing (EMNLP-IJCNLP). Hong Kong, China: Association for Computational Linguistics, 2019: 2567-2577.

[14] Pal A, Umapathi L K, Sankarasubbu M. MedMCQA: A large-scale multi-subject multi-choice dataset for medical domain question answering[C]//Conference on Health, Inference, and Learning. PMLR, 2022: 248-260.

[15] Singhal K, Azizi S, Tu T, et al. Large language models encode clinical knowledge[J]. Nature, 2023, 620(7972): 172-180.

[16] Guha N, Nyarko J, Ho D, et al. LegalBench: A collaboratively built benchmark for measuring legal reasoning in large language models[J]. SSRN Electronic Journal, 2023, 36: 44123-44279.

[17] Guo X, Xia H, Liu Z, et al. FinEval: A Chinese financial domain knowledge evaluation benchmark for large language models[C]//Proceedings of the 2025 Conference of the Nations of the Americas Chapter of the Association for Computational Linguistics: Human Language Technologies (Volume 1: Long Papers). Albuquerque, New Mexico, 2025: 6258-6292.

[18] Li M, Zhong J, Chen T, et al. EEE-Bench: A comprehensive multimodal electrical and electronics engineering benchmark[C]//2025 IEEE/CVF Conference on Computer Vision and Pattern Recognition (CVPR). Nashville, TN, USA, 2025: 13337-13349.

[19] Zhang S, Huang Q, Zhang Q, et al. ElecBench: A large language model benchmark in electric power domain[J]. Engineering Applications of Artificial Intelligence, 2025, 162: 112310.